\documentclass{llncs}

\usepackage{wrapfig}

\usepackage{graphicx}
\usepackage{epsfig}
\usepackage{verbatim}

\usepackage{algorithm,algorithmic}
\usepackage{url,setspace, floatflt}
\usepackage{amssymb}

\urldef{\mailsa}\path|costapan@eng.auth.gr,  petroula@gen.auth.gr|

\begin{document}
\title{The Perceptron with Dynamic Margin}
\author{Constantinos Panagiotakopoulos \and Petroula Tsampouka}
\institute{Physics Division, School of Technology\\ Aristotle University of Thessaloniki, Greece \\
\mailsa\\}
\maketitle
\begin{abstract}
The classical perceptron rule provides a varying upper bound on the maximum margin, namely the length of the current weight vector divided by the total number of updates up to that time. Requiring that the perceptron updates its internal state whenever the normalized margin of a pattern is found not to exceed a certain fraction of this dynamic upper bound we construct a new approximate maximum margin classifier called the perceptron with dynamic margin (PDM). We demonstrate that PDM converges in a finite number of steps and derive an upper bound on them. We also compare experimentally PDM with other perceptron-like algorithms and support vector machines on hard margin tasks involving linear kernels which are equivalent to 2-norm soft margin.\\

\end{abstract}

\renewcommand{\vec}[1]{\mbox{\boldmath$#1$}}
\newcommand{\Tiny}[1]{\mbox{\tiny$#1$}}

\section{Introduction}
It is a common belief that learning machines able to produce solution hyperplanes with large margins exhibit greater generalization ability \cite{Vap} and this justifies the enormous interest in Support Vector Machines (SVMs) \cite{Vap,CST}. Typically, SVMs obtain large margin solutions by solving a constrained quadratic optimization problem using dual variables. In their native form, however, efficient implementation is hindered by the quadratic dependence of their memory requirements in the number of training examples a fact which renders prohibitive the processing of large datasets. To overcome this problem decomposition methods \cite{Plat,Joa} were developed that apply optimization only to a subset of the training set. Although such methods led to improved convergence rates, in practice their superlinear dependence on the number of examples, which can be even cubic, can still lead to excessive runtimes when large datasets are processed. Recently, the so-called linear SVMs \cite{Joa06,HCL,PT2} made their appearance. They take advantage of linear kernels in order to allow parts of them to be written in primal notation and were shown to outperform decomposition SVMs when dealing with massive datasets.

The above considerations motivated research in alternative large margin classifiers naturally formulated in primal space long before the advent of linear SVMs. Such algorithms are mostly based on the perceptron \cite{Ros,Nov}, the simplest online learning algorithm for binary linear classification. Like the perceptron, they focus on the primal problem by updating a weight vector which represents at each step the current state of the algorithm whenever a data point presented to it satisfies a specific condition. It is the ability of such algorithms to process one example at a time\footnote{The conversion of online algorithms to the batch setting is done by cycling repeatedly through the dataset and using the last hypothesis for prediction.} that allows them to spare time and memory resources and consequently makes them able to handle large datasets. The first algorithm of that kind is the perceptron with margin \cite{DH} which is much older than SVMs. It is an immediate extension of the perceptron which provably achieves solutions with only up to $1/2$ of the maximum margin \cite{KM}. Subsequently, various algorithms succeeded in approximately attaining maximum margin by employing modified perceptron-like update rules. Such algorithms include ROMMA \cite{LL}, ALMA \cite{Gen}, CRAMMA \cite{TST1} and MICRA \cite{TST2}. Very recently, the same goal was accomplished by a generalized perceptron with margin, the margitron \cite{PT1}.

The most straightforward way of obtaining large margin solutions through a perceptron is by requiring that the weight vector be updated every time the example presented to the algorithm has (normalized) margin which does not exceed a predefined value \cite{TST0,TST,Blum}. The obvious problem with this idea, however, is that the algorithm with such a fixed margin condition will definitely not converge unless the target value of the margin is smaller than the unknown maximum margin. In an earlier work \cite{PT1} we noticed that the upper bound $\left\| \vec a_t \right\|/t$ on the maximum margin, with $\left\| \vec a_t \right\|$ being the length of the weight vector and $t$ the number of updates, that comes as an immediate consequence of the perceptron update rule is very accurate and tends to improve as the algorithm achieves larger margins. In the present work we replace the fixed target margin value with a fraction $1-\epsilon$ of this varying upper bound on the maximum margin. The hope is that as the algorithm keeps updating its state the upper bound will keep approaching the maximum margin and convergence to a solution with the desired accuracy $\epsilon$ will eventually occur. Thus, the resulting algorithm may be regarded as a realizable implementation of the perceptron with fixed margin condition.

The rest of this paper is organized as follows. Section 2 contains some preliminaries and a motivation of the algorithm based on a qualitative analysis. In Sect. 3 we give a formal theoretical analysis. Section 4 is devoted to implementational issues. Section 5 contains our experimental results while Sect. 6 our conclusions.

\section{Motivation of the Algorithm}
Let us consider a linearly separable training set $\{(\vec x_k, l_k)\}^m_{k=1}$, with vectors $\vec x_k\in \bbbr^d$ and labels $l_k \in \{+1,-1\}$. This training set may either be the original dataset or the result of a mapping into a feature space of higher dimensionality \cite{Vap,CST}. Actually, there is a very well-known construction \cite{FS} making linear separability always possible, which amounts to the adoption of the 2-norm soft margin. By placing $\vec x_k$ in the same position at a distance $\rho$ in an additional dimension, i.e. by extending $\vec x_k$ to $[\vec x_k, \rho]$, we construct an embedding of our data into the so-called augmented space \cite{DH}. This way, we construct hyperplanes possessing bias in the non-augmented feature space. Following the augmentation, a reflection with respect to the origin of the negatively labeled patterns is performed by multiplying every pattern with its label. This allows for a uniform treatment of both categories of patterns.  Also, $R\equiv\displaystyle \max_{k} \left\| \vec{y}_{k} \right\|$ with $\vec{y}_{k} \equiv [l_k\vec x_k, l_k\rho]$ the $k^{\rm {th}}$ augmented and reflected pattern. Obviously, $R \ge \rho $. 

The relation characterizing optimally correct classification of the training patterns $\vec{y}_{k}$ by a 
weight vector $\vec{u}$ of unit norm in the augmented space is
\begin{equation}
\label{gamma}
\vec{u} \cdot \vec{y}_{k}\ge \gamma_{\rm d}\equiv \displaystyle \max_{\Tiny{ \vec{u}^{\prime}:\left\|\vec{u}^{\prime}\right\|=1}}
\displaystyle \min_{i}\left \{\vec{u^{\prime}} \cdot \vec{y}_{i}\right \}  \ \ \ \ \forall k \enspace.
\end{equation}
We shall refer to $\gamma_{\rm d}$ as the maximum directional margin. It coincides with
the maximum margin in the augmented space with respect to hyperplanes passing through
the origin. For the maximum directional margin $\gamma_{\rm d}$ and the maximum geometric margin $\gamma$ in the non-augmented feature space, it holds that $1\le {\gamma}/{\gamma_{\rm d}}\le {R}/{\rho}$.
As $\rho \to \infty$, ${R}/{\rho}\to 1$ and, consequently, $\gamma_{\rm d} \to \gamma$ \cite{TST0,TST}.  

We consider algorithms in which the augmented weight vector $\vec{a}_{t}$ is initially set to zero, i.e. $\vec{a}_{0}=\vec0$, and is updated according to the classical perceptron rule
\begin{equation}
\label{update}
\vec{a}_{t+1}=\vec{a}_{t}+\vec{y}_k
\end{equation}
each time an appropriate misclassification condition is satisfied by a training pattern $\vec{y}_{k}$. Taking the inner product of (\ref{update}) with the optimal direction $\vec u$ and using (\ref{gamma}) we get
\[
\vec u \cdot \vec a_{t+1} -\vec u \cdot \vec a_{t} =\vec u \cdot \vec y_k \ge \gamma_{\rm d}
\]
a repeated application of which gives \cite{Nov} 
\begin{equation}
\label{lbound}
\left\| \vec a_t \right\| \ge \vec u \cdot \vec a_t \ge \gamma_{\rm d}t \enspace.
\end{equation}
From (\ref{lbound}) we readily obtain
\begin{equation}
\label{lbound1}
\gamma_{\rm d} \le \frac{\left\| \vec a_t \right\|}{t}
\end{equation}
provided $t>0$. Notice that the above upper bound on the maximum directional margin $\gamma_{\rm d}$ is an immediate consequence of the classical perceptron rule and holds independent of the misclassification condition.

It would be very desirable that $\left\| \vec a_t \right\|/t$ approaches $\gamma_{\rm d}$ with $t$ increasing since this would provide an after-run estimate of the accuracy achieved by an algorithm employing the classical perceptron update. More specifically, with $\gamma^{\prime}_{\rm d}$ being the directional margin achieved upon convergence of the algorithm in $t_{\rm c}$ updates, it holds that
\begin{equation}
\label{est}
\frac{\gamma_{\rm d}-\gamma^{\prime}_{\rm d}}{\gamma_{\rm d}} \le 1-\frac{\gamma^{\prime}_{\rm d}t_{\rm c}}{\left\| \vec a_{t_{\rm c}} \right\|} \enspace.
\end{equation}

In order to understand the mechanism by which $\left\| \vec a_t \right\|/t$ evolves we consider the difference between two consecutive values of $\left\| \vec a_t \right\|^2/t^2$ which may be shown to be given by the relation
\begin{equation}
\label{dif}
\frac{\left\|\vec{a_t}\right\|^2}{t^2}-\frac{\left\|\vec{a_{t+1}}\right\|^2}{(t+1)^2}=\frac{1}{t(t+1)}
\left\{\left(\frac{\left\|\vec{a_t}\right\|^2}{t}-\vec a_t \cdot \vec y_k \right)+ \left(\frac{\left\|\vec{a_{t+1}}\right\|^2}{t+1}-\vec a_{t+1} \cdot \vec y_k \right)\right\} \enspace.
\end{equation}
Let us assume that satisfaction of the misclassification condition by a pattern $\vec{y}_{k}$ has as a consequence that ${\left\|\vec{a_t}\right\|^2}/{t}>\vec a_t \cdot \vec y_k$ (i.e., the normalized margin $\vec u_t \cdot \vec y_k$ of $\vec y_k$ (with $\vec u_t \equiv \vec a_t/\left\|\vec{a_t}\right\|$) is smaller than the upper bound (\ref{lbound1}) on $\gamma_{\rm d}$). Let us further assume that after the update has taken place $\vec{y}_{k}$ still satisfies the misclassification condition and therefore ${\left\|\vec{a_{t+1}}\right\|^2}/({t+1})>\vec a_{t+1} \cdot \vec y_k $. Then, the r.h.s. of (\ref{dif}) is positive and $\left\| \vec a_t \right\|/t$ decreases as a result of the update. In the event, instead, that the update leads to violation of the misclassification condition, ${\left\|\vec{a_{t+1}}\right\|^2}/({t+1})$ is not necessarily larger than $\vec a_{t+1} \cdot \vec y_k $ and $\left\| \vec a_t \right\|/t$ may not decrease as a result of the update. We expect that statistically, at least in the early stages of the algorithm, most updates do not lead to correctly classified patterns (i.e., patterns which violate the misclassification condition) and as a consequence $\left\| \vec a_t \right\|/t$ will have the tendency to decrease. Obviously, the rate at which this will take place depends on the size of the difference $\left\|\vec{a_t}\right\|^2/{t}-\vec a_t \cdot \vec y_k$ which, in turn, depends on the misclassification condition.

If we are interested in obtaining solutions possessing margin the most natural choice of misclassification condition is the fixed (normalized) margin condition 
\begin{equation}
\label{fixed2}
\vec a_t \cdot \vec y_k \le (1-\epsilon)\gamma_{\rm d}\left\|\vec{a_t}\right\|
\end{equation}
with the accuracy parameter $\epsilon$ satisfying $0<\epsilon \le 1$. This is an example of a misclassification condition which if it is satisfied ensures that ${\left\|\vec{a_t}\right\|^2}/{t}>\vec a_t \cdot \vec y_k$. Moreover, by making use of (\ref{lbound1}) and (\ref{fixed2}) it may easily be shown that ${\left\|\vec{a_{t+1}}\right\|^2}/({t+1}) \ge \vec a_{t+1} \cdot \vec y_k$ for $t \ge \epsilon^{-1}R^2/\gamma^2_{\rm d}$. Thus, after at most $\epsilon^{-1}R^2/\gamma^2_{\rm d}$ updates $\left\| \vec a_t \right\|/t$ decreases monotonically. The perceptron algorithm with fixed margin condition (PFM) is known to converge in a finite number of updates to an $\epsilon$-accurate approximation of the maximum directional margin hyperplane \cite{TST0,TST,Blum}. Although it appears that PFM demands exact knowledge of the value of $\gamma_{\rm d}$, we notice that only the value of
$\beta\equiv(1-\epsilon)\gamma_{\rm d}$, which is the quantity entering (\ref{fixed2}), needs to be set and not the values of $\epsilon$ and $\gamma_{\rm d}$ separately. That is why the after-run estimate (\ref{est}) is useful in connection with the algorithm in question. Nevertheless, in order to make sure that $\beta <\gamma_{\rm d}$ a priori knowledge of a fairly good lower bound on $\gamma_{\rm d}$ is required and this is an obvious defect of PFM.

The above difficulty associated with the fixed margin condition may be remedied if the unknown $\gamma_{\rm d}$ is replaced for $t>0$ with its varying upper bound $\left\| \vec a_t \right\|/t$    
\begin{equation}
\label{dyn}
\vec a_t \cdot \vec y_k \le (1-\epsilon)\frac{\left\|\vec{a_t}\right\|^2}{t} \enspace.
\end{equation}
Condition (\ref{dyn}) ensures that ${\left\|\vec{a_t}\right\|^2}/{t}-\vec a_t \cdot \vec y_k \ge \epsilon {\left\|\vec{a_t}\right\|^2}/{t}>0$. Moreover, as in the case of the fixed margin condition, ${\left\|\vec{a_{t+1}}\right\|^2}/({t+1})-\vec a_{t+1}\cdot\vec y_k \ge 0$ for $t \ge \epsilon^{-1}R^2/\gamma^2_{\rm d}$. As a result, after at most $\epsilon^{-1}R^2/\gamma^2_{\rm d}$ updates the r.h.s. of (\ref{dif}) is bounded from below by $\epsilon \left\| \vec a_t \right\|^2/t^2(t+1) \ge \epsilon\gamma^2_{\rm d}/(t+1)$ and $\left\| \vec a_t \right\|/t$ decreases monotonically and sufficiently fast. Thus, we expect that $\left\| \vec a_t \right\|/t$ will eventually approach $\gamma_{\rm d}$ close enough, thereby allowing for convergence of the algorithm to an $\epsilon$-accurate approximation of the maximum directional margin hyperplane. It is also apparent that the decrease of $\left\| \vec a_t \right\|/t$ will be faster for larger values of $\epsilon$.
\begin{wrapfigure}{l}{0.55\textwidth}
\vspace{-0pt}
\epsfig{file=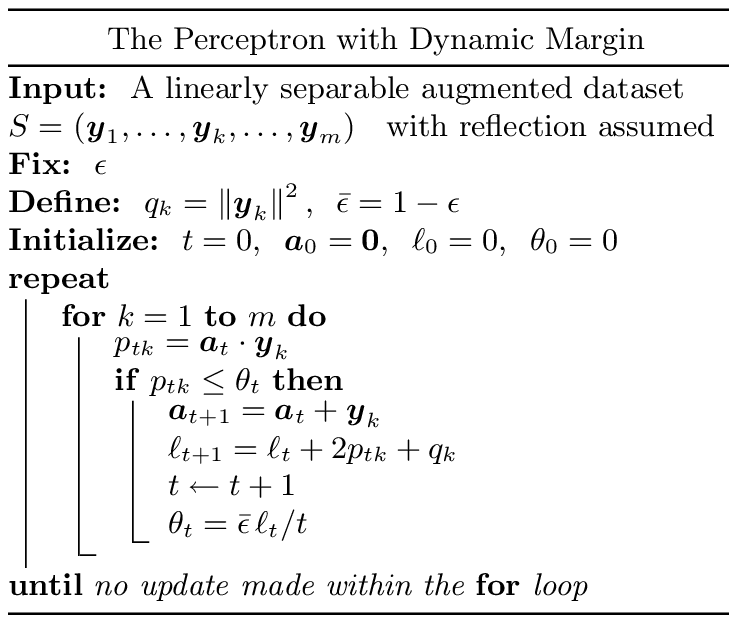, width=0.54\textwidth}
\vspace{-20pt}
\end{wrapfigure}
The perceptron algorithm employing the misclassification condition (\ref{dyn}) (with its threshold set to $0$ for $t=0$), which may be regarded as originating from (\ref{fixed2}) with $\gamma_{\rm d}$ replaced for $t>0$ by its dynamic upper bound $\left\| \vec a_t \right\|/t$, will be named the perceptron with dynamic margin (PDM).

\section{Theoretical Analysis}
From the discussion that led to the formulation of PDM it is apparent that if the algorithm converges it will achieve by construction a solution possessing directional margin at least as large as $(1-\epsilon)\gamma_{\rm d}$. (We remind the reader that convergence assumes violation of the misclassification condition (\ref{dyn}) by all patterns. In addition, (\ref{lbound1}) holds.) The same obviously applies to PFM. Thus, for both algorithms it only remains to be demonstrated that they converge in a finite number of steps. This has already been shown for PFM \cite{TST0,TST,Blum} but no general $\epsilon$-dependent bound in closed form has been derived. Our purpose in this section is to demonstrate convergence of PDM and provide explicit bounds for both algorithms. 

Before we proceed with our analysis we will need the following result.

\begin{lemma}
\label{lemma}
Let the variable $t \ge e^{-C}$ satisfy the inequality
\begin{equation}
\label{lem1}
t < \delta (1+C+\ln t) \enspace,
\end{equation}
where $\delta$, $C$ are constants and $\delta > e^{-C}$. Then
\begin{equation}
\label{lem2}
 t \le t_0 \equiv (1+e^{-1}) \delta \left(C+ \ln \left((1+e) \delta\right) \right) \enspace.
\end{equation}
\end{lemma}
 
\begin{proof}
If $t \ge e^{-C}$ then $(1+C+ \ln t) \ge 1$ and inequality (\ref{lem1}) is equivalent to $f(t)=t/(1+C+ \ln t) - \delta <0$. For the function $f(t)$ defined in the interval $[e^{-C}, +\infty)$ it holds that $f(e^{-C})<0$ and $df/dt=(C+ \ln t)/(1+C+ \ln t)^2 >0$ for $t>e^{-C}$. Stated differently, $f(t)$ starts from negative values at $t=e^{-C}$ and increases monotonically. Therefore, if $f(t_0) \ge 0$ then $t_0$ is an upper bound of all $t$ for which $f(t)<0$. Indeed, it is not difficult to verify that $t_0>\delta>e^{-C}$ and
\[
f(t_0)=\delta \left( {(1+e^{-1})}\left({1+\frac{\ln \ln (e^C(1+e)\delta)}{\ln (e^C(1+e)\delta)}}\right)^{-1}-1\right) \ge 0
\]
given that $\ln \ln x / \ln x \le e^{-1}$. \qed
\end{proof}

Now we are ready to derive an upper bound on the number of steps of PFM.

\begin{theorem}
\label{theorem1}
The number $t$ of updates of the perceptron algorithm with fixed margin condition satisfies the bound
\[
t \le \frac{(1+e^{-1})}{2\epsilon }\frac{R^2}{\gamma^2_{\rm d}}\left\{4\frac{\gamma_{\rm d}}{R}\left(1-\frac{\gamma_{\rm d}}{R}(1-\epsilon)\right)+\ln \left(\frac{(1+e)}{\epsilon}\frac{R}{\gamma_{\rm d}}\left(1-\frac{\gamma_{\rm d}}{R}(1-\epsilon) \right) \right) \right\} \enspace.
\]
\end{theorem}

\begin{proof}
From (\ref{update}) and (\ref{fixed2}) we get
\[
\left\|\vec a_{t+1}\right\|^2  = \left\|\vec{a_t}\right\|^2+\left\|\vec{y_k}\right\|^2+2\vec a_t \cdot \vec y_k  \le  
 \left\|\vec a_t\right\|^2 \left(1+\frac{R^2}{\left\|\vec a_t\right\|^2 }+\frac{2(1-\epsilon)\gamma_{\rm d}}{\left\|\vec a_t\right\|}\right) \enspace.
\]
Then, taking the square root and using the inequality $\sqrt{1+x} \le 1+x/2$ we have
\[
\left\|\vec a_{t+1}\right\| \le \left\|\vec a_t\right\|\left(1+\frac{R^2}{\left\|\vec a_t\right\|^2 }+\frac{2(1-\epsilon)\gamma_{\rm d}}{\left\|\vec a_t\right\|}\right)^{\frac{1}{2}} \le \left\|\vec a_t\right\| \left(1+\frac{R^2}{2\left\|\vec a_t\right\|^2 }+\frac{(1-\epsilon)\gamma_{\rm d}}{\left\|\vec a_t\right\|}\right) \enspace.
\]
Now, by making use of $\left\|\vec a_t\right\| \ge \gamma_{\rm d}t$, we observe that
\[
\left\|\vec a_{t+1}\right\|-\left\|\vec a_t\right\| \le \frac{R^2}{2\left\|\vec a_t\right\|}+(1-\epsilon)\gamma_{\rm d} \le \frac{R^2}{2\gamma_{\rm d}}\frac{1}{t}+(1-\epsilon)\gamma_{\rm d} \enspace.
\]
A repeated application of the above inequality $t-N$ times ($t>N \ge 1$) gives
\begin{eqnarray}
\left\|\vec a_{t}\right\|-\left\|\vec a_N\right\| & \le & \frac{R^2}{2\gamma_{\rm d}}\sum^{t-1}_{k=N}k^{-1}+(1-\epsilon)\gamma_{\rm d}(t-N) \nonumber \\
& < & \frac{R^2}{2\gamma_{\rm d}}\left(\frac{1}{N}+\int^t_N k^{-1}dk\right)+(1-\epsilon)\gamma_{\rm d}(t-N) \nonumber
\end{eqnarray}
from where using the obvious bound $\left\|\vec a_{N}\right\| \le RN$ we get an upper bound on $\left\|\vec a_{t}\right\|$
\[
\left\|\vec a_{t}\right\|<\frac{R^2}{2\gamma_{\rm d}}\left(\frac{1}{N}+\ln \frac{t}{N}\right)+(1-\epsilon)\gamma_{\rm d}(t-N)+RN \enspace.
\]
Combining the above upper bound on $\left\|\vec a_{t}\right\|$, which holds not only for $t>N$ but also for $t=N$, with the lower bound from (\ref{lbound}) we obtain
\[
t<\frac{1}{2\epsilon}\frac{R^2}{\gamma^2_{\rm d}}\left\{\frac{1}{N}- \ln N +2\frac{\gamma_{\rm d}}{R}\left(1-\frac{\gamma_{\rm d}}{R}(1-\epsilon)\right)N +\ln t  \right\} \enspace.
\]
Setting
\[
\delta= \frac{1}{2\epsilon}\frac{R^2}{\gamma^2_{\rm d}}\;, \;\;\;\;\;\; \alpha=2\frac{\gamma_{\rm d}}{R}\left(1-\frac{\gamma_{\rm d}}{R}(1-\epsilon)\right)
\]
and choosing $N=1+[\alpha^{-1}]$, with $[x]$ being the integer part of $x \ge 0$, we finally get  
\begin{equation}
\label{tboundfx}
t<\delta(1+2\alpha+\ln \alpha+\ln t) \enspace.
\end{equation}
Notice that in deriving (\ref{tboundfx}) we made use of the fact that $\alpha N+N^{-1}-\ln N < 1+2\alpha+\ln \alpha$.
Inequality (\ref{tboundfx}) has the form (\ref{lem1}) with $C=2\alpha+\ln \alpha$. Obviously, $e^{-C}<\alpha^{-1}<N \le t$ and $e^{-C}<\alpha^{-1} \le \delta$. Thus, the conditions of Lemma \ref{lemma} are satisfied and the required bound, which is of the form (\ref{lem2}), follows from (\ref{tboundfx}). \qed
\end{proof}

Finally, we arrive at our main result which is the proof of convergence of PDM in a finite number of steps and the derivation of the relevant upper bound.

\begin{theorem}
\label{theorem2}
The number $t$ of updates of the perceptron algorithm with dynamic margin satisfies the bound

\[
\hspace{-6cm}t \le \left\{ 
\begin{minipage}[b]{5cm}

\vspace{-10pt}
\[
\begin{array}{l l c c}
t_0 \left(1-\frac{1}{1-2\epsilon}\frac{R^2}{\gamma^2_{\rm d}}t^{-1}_0 \right)^{\frac{1}{2\epsilon}},\, &
t_0 \equiv [\epsilon^{-1}]\left(\frac{R}{\gamma_{\rm d}}\right)^{\frac{1}{\epsilon}}
\left(1+ \frac{[\epsilon^{-1}]^{-1}}{1-2\epsilon}\right)^{\frac{1}{2\epsilon}} \;\; & \rm {if}\,\, \epsilon<\frac{1}{2}\\ \\
(1+e^{-1})\frac{R^2}{\gamma^2_{\rm d}}\ln \left( (1+e)\frac{R^2}{\gamma^2_{\rm d}}\right) & & \rm {if}\,\, \epsilon=\frac{1}{2} \\ \\
t_0\left( 1-2(1-\epsilon)t_0^{1-2\epsilon}\right),\, & t_0 \equiv \frac{\epsilon(3-2\epsilon)}{2\epsilon-1}\frac{R^2}{\gamma^2_{\rm d}}\;\; & \rm {if}\,\, \epsilon>\frac{1}{2} & .
\end{array}
\]
\end{minipage}
 \right.
\]

\end{theorem}

\begin{proof}
From (\ref{update}) and (\ref{dyn}) we get
\begin{equation}
\label{eq1}
\left\|\vec a_{t+1}\right\|^2  = \left\|\vec{a_t}\right\|^2+2\vec a_t \cdot \vec y_k +\left\|\vec{y_k}\right\|^2 \le  
 \left\|\vec a_t\right\|^2 \left(1+\frac{2(1-\epsilon)}{t}\right)+R^2 \enspace.
\end{equation}

Let us assume that $\epsilon < 1/2$. Then, using the inequality $(1+x)^{\zeta} \ge 1+\zeta x$ for $x \ge 0$, $\zeta=2(1-\epsilon) \ge 1$ in (\ref{eq1}) we obtain
\[
\left\|\vec a_{t+1}\right\|^2 \le \left\|\vec a_t\right\|^2\left(1+\frac{1}{t}\right)^{2(1-\epsilon)}+R^2
\]
from where by dividing both sides with $(t+1)^{2(1-\epsilon)}$ we arrive at
\[
\frac{\left\|\vec a_{t+1}\right\|^2}{(t+1)^{2(1-\epsilon)}}-\frac{\left\|\vec a_t\right\|^2}{t^{2(1-\epsilon)}}
\le \frac{R^2}{(t+1)^{2(1-\epsilon)}} \enspace.
\]
A repeated application of the above inequality $t-N$ times ($t>N \ge 1$) gives
\begin{eqnarray}
\label{eq2}
\frac{\left\|\vec a_t \right\|^2}{t^{2(1-\epsilon)}}-\frac{\left\|\vec a_N\right\|^2}{N^{2(1-\epsilon)}}
\le R^2\sum^t_{k=N+1}k^{-2(1-\epsilon)} & \le & R^2\int^t_N k^{-2(1-\epsilon)}dk \nonumber \\
& = & \frac{R^2 N^{2\epsilon-1}}{2\epsilon-1}\left( \left( \frac{t}{N}\right)^{2\epsilon-1}-1\right).
\end{eqnarray}
Now, let us define
\[
\alpha_t \equiv \frac{\left\|\vec a_t \right\|}{Rt}
\]
and observe that the bounds $\left\|\vec a_t \right\| \le Rt$ and $\left\|\vec a_t \right\| \ge \gamma_{\rm d}t$ confine $\alpha_t$ to lie in the range
\[
\frac{\gamma_{\rm d}}{R} \le \alpha_t \le 1 \enspace. 
\]
Setting $\left\|\vec a_N \right\|=\alpha_N RN$ in (\ref{eq2}) we get the following upper bound on $\left\|\vec a_t \right\|^2$
\[
\left\|\vec a_t \right\|^2 \le t^{2(1-\epsilon)}\alpha^2_NR^2N^{2\epsilon}\left\{1+\frac{\alpha^{-2}_N N^{-1}}{2\epsilon-1}\left( \left( \frac{t}{N}\right)^{2\epsilon-1}-1\right) \right\}
\]
which combined with the lower bound $\left\|\vec a_t \right\|^2 \ge \gamma^2_{\rm d}t^2$ leads to
\begin{equation}
\label{boundt1}
t^{2\epsilon} \le \alpha^2_N\frac{R^2}{\gamma^2_{\rm d}}N^{2\epsilon}\left\{1+\frac{\alpha^{-2}_N N^{-1}}{2\epsilon-1}\left( \left( \frac{t}{N}\right)^{2\epsilon-1}-1\right) \right\} \enspace.
\end{equation}
For $\epsilon < 1/2$ the term proportional to $(t/N)^{2\epsilon-1}$ in (\ref{boundt1}) is negative and may be dropped to a first approximation leading to the looser upper bound $t_0$
\begin{equation}
\label{boundt01}
t_0 \equiv N \left(\alpha_N\frac{R}{\gamma_{\rm d}}\right)^{\frac{1}{\epsilon}}\left(1+ \frac{\alpha^{-2}_N N^{-1}}{1-2\epsilon}\right)^{\frac{1}{2\epsilon}}
\end{equation}
on the number $t$ of updates. Then, we may replace $t$ with its upper bound $t_0$ in the r.h.s. of (\ref{boundt1}) and get the improved bound
\[
t \le t_0 \left(1-\frac{1}{1-2\epsilon}\frac{R^2}{\gamma^2_{\rm d}}t^{-1}_0 \right)^{\frac{1}{2\epsilon}}\enspace.
\]
This is allowed given that the term proportional to $(t/N)^{2\epsilon-1}$ in (\ref{boundt1}) is negative and moreover $t$ is raised to a negative power. Choosing $N=[\epsilon^{-1}]$ and $\alpha_N=1$ (i.e., setting $\alpha_N$ to its upper bound which is the least favorable assumption) we obtain the bound stated in Theorem \ref{theorem2} for $\epsilon < 1/2$.

Now, let $\epsilon > 1/2$. Then, using the inequality $(1+x)^{\zeta}+\zeta(1-\zeta)x^2/2 \ge 1+\zeta x$ for $x \ge 0$, $0 \le \zeta=2(1-\epsilon) \le 1$ in (\ref{eq1}) and the bound $\left\|\vec a_t\right\| \le Rt$ we obtain
\begin{eqnarray}
\left\|\vec a_{t+1}\right\|^2 & \le & \left\|\vec a_t\right\|^2\left(1+\frac{1}{t}\right)^{2(1-\epsilon)}+
(1-\epsilon)(2\epsilon-1)\frac{\left\|\vec a_t\right\|^2}{t^2}+R^2 \nonumber \\
& \le & \left\|\vec a_t\right\|^2\left(1+\frac{1}{t}\right)^{2(1-\epsilon)}+\epsilon(3-2\epsilon)R^2 \nonumber \enspace.
\end{eqnarray}
By dividing both sides of the above inequality with $(t+1)^{2(1-\epsilon)}$ we arrive at
\begin{equation}
\frac{\left\|\vec a_{t+1}\right\|^2}{(t+1)^{2(1-\epsilon)}}-\frac{\left\|\vec a_t\right\|^2}{t^{2(1-\epsilon)}}
\le \epsilon(3-2\epsilon)\frac{R^2}{(t+1)^{2(1-\epsilon)}}
\end{equation}
a repeated application of which, using also $\left\|\vec a_1 \right\|^2\le R^2 \le \epsilon(3-2\epsilon)R^2$, gives
\begin{eqnarray}
\frac{\left\|\vec a_t \right\|^2}{t^{2(1-\epsilon)}}
\le \epsilon(3-2\epsilon)R^2\sum^t_{k=1}k^{-2(1-\epsilon)} & \le & \epsilon(3-2\epsilon)R^2\left(1+\int^t_1 k^{-2(1-\epsilon)}dk \right) \nonumber \\
& = & \epsilon(3-2\epsilon)R^2 \left(1+\frac{t^{2\epsilon-1}-1}{2\epsilon-1}\right)\nonumber\enspace.
\end{eqnarray}
Combining the above bound with the bound $\left\|\vec a_t \right\|^2 \ge \gamma^2_{\rm d}t^2$ we obtain
\begin{equation}
\label{boundt2}
t^{2\epsilon} \le \epsilon(3-2\epsilon)\frac{R^2}{\gamma^2_{\rm d}}\left(1+\frac{t^{2\epsilon-1}-1}{2\epsilon-1}  \right)
\end{equation}
or
\begin{equation}
\label{boundt3}
t \le \frac{\epsilon(3-2\epsilon)}{2\epsilon-1}\frac{R^2}{\gamma^2_{\rm d}}\left(1-2(1-\epsilon)t^{1-2\epsilon}\right)\enspace.
\end{equation}
For $\epsilon > 1/2$ the term proportional to $t^{1-2\epsilon}$ in (\ref{boundt3}) is negative and may be dropped to a first approximation leading to the looser upper bound $t_0$
\[
t_0 \equiv \frac{\epsilon(3-2\epsilon)}{2\epsilon-1}\frac{R^2}{\gamma^2_{\rm d}}
\]
on the number $t$ of updates. Then, we may replace $t$ with its upper bound $t_0$ in the r.h.s. of (\ref{boundt3}) and get the improved bound stated in Theorem \ref{theorem2} for $\epsilon > 1/2$.
This is allowed given that the term proportional to $t^{1-2\epsilon}$ in (\ref{boundt3}) is negative and moreover $t$ is raised to a negative power. 

Finally, taking the limit $\epsilon \to 1/2$ in (\ref{boundt1}) (with $N=1$, $\alpha_N=1$) or in (\ref{boundt2}) we get 
\[
t\le \frac{R^2}{\gamma^2_{\rm d}}\left(1+ \ln t \right)
\]
which on account of Lemma \ref{lemma} leads to the bound of Theorem \ref{theorem2} for $\epsilon = 1/2$. \qed
\end{proof}

\begin{remark}
The bound of Theorem \ref{theorem2} holds for PFM as well on account of (\ref{lbound1}).
\end{remark}

The worst-case bound of Theorem \ref{theorem2} for $\epsilon \ll 1$ behaves like $\epsilon^{-1}({R}/{\gamma_{\rm d}})^{\frac{1}{\epsilon}}$ which suggests an extremely slow convergence if we require margins close to the maximum. From expression (\ref{boundt01}) for $t_0$, however, it becomes apparent that a more favorable assumption concerning the value of $\alpha_N$ (e.g., $\alpha_N \ll 1$ or even as low as $\alpha_N \sim \gamma_{\rm d}/R$) after the first $N \gg \alpha^{-2}_N$ updates does lead to tremendous improvement provided, of course, that $N$ is not extremely large. Such a sharp decrease of $\left\| \vec a_t \right\|/t$ in the early stages of the algorithm, which may be expected from relation (\ref{dif}) and the discussion that followed, lies behind its experimentally exhibited rather fast convergence.

It would be interesting to find a procedure by which the algorithm will be forced to a guaranteed sharp decrease of the ratio $\left\| \vec a_t \right\|/t$. The following two observations will be vital in devising such a procedure. First, we notice that when PDM with accuracy parameter $\epsilon$ has converged in $t_{\rm c}$ updates the threshold $(1-\epsilon){\left\|\vec{a_{t_{\rm c}}}\right\|^2}/{t_{\rm c}}$ of the misclassification condition must have fallen below $\gamma_{\rm d}\left\|\vec{a_{t_{\rm c}}}\right\|$. Otherwise, the normalized margin $\vec u_{t_{\rm c}} \cdot \vec y_k$ of all patterns $\vec y_k$ would be larger than $\gamma_{\rm d}$. Thus, $\alpha_{t_{\rm c}} < (1-\epsilon)^{-1}\gamma_{\rm d}/R$. Second, after convergence of the algorithm with accuracy parameter $\epsilon_1$ in $t_{{\rm c}_1}$ updates we may lower the accuracy parameter from the value $\epsilon_1$ to the value $\epsilon_2$ and continue the run from the point where convergence with parameter $\epsilon_1$ has taken place since for all updates that took place during the first run the misclassified patterns would certainly satisfy (at that time) the condition associated with the smaller parameter $\epsilon_2$. This way, the first run is legitimately fully incorporated into the second one and the $t_{{\rm c}_1}$ updates required for convergence during the first run may be considered the first $t_{{\rm c}_1}$ updates of the second run under this specific policy of presenting patterns to the algorithm. Combining the above two observations we see that by employing a first run with accuracy parameter $\epsilon_1$ we force the algorithm with accuracy parameter $\epsilon_2<\epsilon_1$ to have $\alpha_t$ decreased from a value $\sim 1$ to a value $\alpha_{t_{{\rm c}_1}} < (1-\epsilon_1)^{-1}\gamma_{\rm d}/R$ in the first $t_{{\rm c}_1}$ updates. 

The above discussion suggests that we consider a decreasing sequence of parameters $\epsilon_n$ such that $\epsilon_{n+1}=\epsilon_n/\eta$ ($\eta>1$) starting with $\epsilon_0=1/2$ and ending with the required accuracy $\epsilon$ and perform successive runs of PDM with accuracies $\epsilon_n$ until convergence in $t_{{\rm c}_n}$ updates is reached. According to our earlier discussion $t_{{\rm c}_n}$ includes the updates that led the algorithm to convergence in the current and all previous runs. Moreover, at the end of the run with parameter $\epsilon_n$ we will have ensured that $\alpha_{t_{{\rm c}_n}} < (1-\epsilon_n)^{-1}\gamma_{\rm d}/R$. Therefore, $t_{{\rm c}_{n+1}}$ satisfies $t_{{\rm c}_{n+1}} \le t_0$ or
\[
t_{{\rm c}_{n+1}} \le t_{{\rm c}_n}\left(\frac{1}{1-\epsilon_n} \right)^{{\eta}/{\epsilon_n}}\left(1+\frac{(1-\epsilon_n)^2}{1-2\epsilon_n/\eta} \frac{R^2}{\gamma^2_{\rm d}}t^{-1}_{{\rm c}_n}\right)^{{\eta}/{2\epsilon_n}} \enspace.
\]
This is obtained by substituting in (\ref{boundt01}) the values $\epsilon=\epsilon_{n+1}=\epsilon_n/\eta$, $N=t_{{\rm c}_n}$ and $\alpha_N=(1-\epsilon_n)^{-1}\gamma_{\rm d}/R$ which is the least favorable choice for $\alpha_{t_{{\rm c}_n}}$.
Let us assume that $\epsilon_n \ll 1$ and set $t_{{\rm c}_n} = \xi^{-1}_nR^2/\gamma^2_{\rm d}$ with $\xi_n \ll 1$. Then, $1/({1-\epsilon_n}) ^{{\eta}/{\epsilon_n}} \simeq e^\eta$ and
\[
\left(1+\frac{(1-\epsilon_n)^2}{1-2\epsilon_n/\eta} \frac{R^2}{\gamma^2_{\rm d}}t^{-1}_{{\rm c}_n}\right)^{{\eta}/{2\epsilon_n}} \simeq (1+\xi_n)^{{\eta}/{2\epsilon_n}} \simeq e^{\eta\xi_n/2\epsilon_n}.
\]
For $\xi_n \simeq \epsilon_n$ the term above becomes approximately $e^{\eta/2}$ while for $\xi_n \ll \epsilon_n$ approaches 1. We see that under the assumption that PDM with accuracy parameter $\epsilon_n$ converges in a number of updates $\gg R^2/\gamma^2_{\rm d}$ the ratio $t_{{\rm c}_{n+1}}/t_{{\rm c}_n}$ in the successive run scenario is rather tightly constrained. If, instead, our assumption is not satisfied then convergence of the algorithm is fast anyway. Notice, that the value of $t_{{\rm c}_{n+1}}/t_{{\rm c}_n}$ inferred from the bound of Theorem \ref{theorem2} is $\sim \eta \left({R}/{\gamma_{\rm d}}\right)^{{(\eta-1)}/{\epsilon_{n}}}$ which is extremely large. We conclude that PDM employing the successive run scenario (PDM-succ) potentially converges in a much smaller number of steps.

\section{Efficient Implementation}

To reduce the computational cost involved in running PDM, we extend the procedure of \cite{PT1,PT2} and construct a three-member nested sequence of reduced ``active sets" of data points. As we cycle once through the full dataset, the (largest) first-level active set is formed from the points of the full dataset satisfying $\vec{a}_{t}\cdot\vec{y}_{k}\le c_1 (1-\epsilon) \left\|\vec a_t \right\|^{2}/t$ with $c_1=2.2$. Analogously, the second-level active set is formed as we cycle once through the first-level active set from the points which satisfy $\vec{a}_{t}\cdot\vec{y}_{k}\le c_2 (1-\epsilon) \left\|\vec a_t \right\|^{2}/t$ with $c_2=1.1$. The third-level active set comprises the points that satisfy $\vec{a}_{t}\cdot\vec{y}_{k}\le (1-\epsilon) \left\|\vec a_t \right\|^{2}/t$ as we cycle once through the second-level active set.
The third-level active set is presented repetitively to the algorithm for $N_{\rm ep_3}$ mini-epochs. Then, the second-level active set is presented $N_{\rm ep_2}$ times. During each round involving the second-level set, a new third-level set is constructed and a new cycle of $N_{\rm ep_3}$ passes begins. When the number of  $N_{\rm ep_2}$ cycles involving the second-level set is reached the first-level set becomes active again leading to the population of a new second-level active set. By invoking the first-level set for the $(N_{\rm ep_1}+1)^{\rm th}$ time, we trigger the loading of the full dataset and the procedure starts all over again until no point is found misclassified among the ones comprising the full dataset.
Of course, the $N_{\rm ep_1}$, $N_{\rm ep_2}$ and $N_{\rm ep_3}$ rounds
are not exhausted if no update takes place during a round. In all experiments we choose $N_{\rm ep_1}=9$, $N_{\rm ep_2}=N_{\rm ep_3}=12$.
In addition, every time we make use of the full dataset we actually employ a permuted  instance of it.
Evidently, the whole procedure amounts to a different way of sequentially presenting the patterns to the algorithm and does not affect the applicability of our theoretical analysis.
A completely analogous procedure is followed for PFM.

An additional mechanism providing a substantial improvement of the computational efficiency is the one of performing multiple updates \cite{PT1,PT2} once a data point is presented to the algorithm. It is understood, of course, that in order for a multiple update to be compatible with our theoretical analysis it should be equivalent to a certain number of updates occuring as a result of repeatedly presenting to the algorithm the data point in question.
For PDM when a pattern $\vec{y}_k$ is found to satisfy the misclassification condition (\ref{dyn}) we perform $\lambda=[\mu_+]+1$ updates at once. Here, $\mu_+$ is the smallest non-negative root of the quadratic equation in the variable $\mu$ derivable from the relation $(t+\mu)\vec{a}_{t+\mu}\cdot\vec{y}_{k}-(1-\epsilon)\left\|\vec a_{t+\mu} \right\|^{2}=0$ in which $\vec{a}_{t+\mu}\cdot\vec{y}_{k}=\vec{a}_{t}\cdot\vec{y}_{k}+\mu \left\|\vec y_{k} \right\|^{2}$ and $\left\|\vec a_{t+\mu} \right\|^{2}=\left\|\vec a_{t} \right\|^{2}+2\mu\vec{a}_{t}\cdot\vec{y}_{k}+\mu^2 \left\|\vec y_{k} \right\|^{2}$. Thus, we require that as a result of the multiple update the pattern violates the misclassification condition. Similarly, we perform multiple updates for PFM.

Finally, in the case of PDM (no successive runs) when we perform multiple updates we start doing so after the first full epoch. This way, we avoid the excessive growth of the length of the weight vector due to the contribution to the solution of many aligned patterns in the early stages of the algorithm which hinders the fast decrease of $\left\| \vec a_t \right\|/t$. Moreover, in this scenario when we select the first-level active set as we go through the full dataset for the first time (first full epoch) we found it useful to set $c_1=c_2=1.1$ instead of $c_1=2.2$.  

\section{Experimental Evaluation}
\label{exper}
\newcommand{\shr}[1]{\resizebox{!}{0.20cm}{\mbox{$#1$}}}
\newcommand{\sh}[1]{\resizebox{!}{0.25cm}{\mbox{$#1$}}}

We compare PDM with several other large margin classifiers on the basis of their ability to achieve fast convergence to a certain approximation of the ``optimal" hyperplane in the feature space where the patterns are linearly separable. For linearly separable data the feature space is the initial instance space whereas for inseparable data (which is the case here) a space extended by as many dimensions as the instances is considered where each instance is placed at a distance $\Delta$ from the origin in the corresponding dimension\footnote{$\vec y_k=[ \bar{\vec y}_k, l_k\Delta \delta _{1k}, \dots , l_k \Delta \delta _{mk}]$, where $\delta _{ij}$ is Kronecker's $\delta$ and $\bar{\vec y}_k$ the projection of the $k^{\rm {th}}$ extended instance $\vec y_k$ (multiplied by its label $l_k$) onto the initial instance space. The feature space mapping defined by the extension commutes with a possible augmentation (with parameter $\rho$) in which case $\bar{\vec y}_k=[l_k  \bar{\vec x}_k, l_k \rho]$. Here $\bar{\vec x}_k$ represents the $k^{\rm {th}}$ data point.}
\cite{FS}. This extension generates a margin of at least $\Delta/\sqrt{m}$.
Moreover, its employment relies on the well-known equivalence between the hard margin optimization in the extended space and the soft margin optimization in the initial instance space with objective function $\left\|{\vec w}\right\|^2+ \Delta^{-2} {\sum_i} {{\bar{\xi}}_i}^2$ involving the weight vector $\vec w$ and the 2-norm of the slacks $\bar{\xi}_i$ \cite{CST}. Of course, all algorithms are required to solve identical hard margin problems. 

The datasets we used for training are: the Adult ($m=32561$ instances, $n=123$ attributes) and Web ($m=49749,\, n=300$) UCI datasets as compiled by Platt \cite{Plat}, the training  set of the KDD04 Physics dataset ($m=50000$, $n=70$ after removing the 8 columns containing missing features) obtainable from \url{http://kodiak.cs.cornell.edu/kddcup/datasets.html}, the Real-sim ($m=72309,\, n=20958$), News20 ($m=19996,\, n=1355191$) and Webspam (unigram treatment with $m=350000,\, n=254$) datasets all available at \url{http://www.csie.ntu.edu.tw/~cjlin/libsvmtools/datasets}, the multiclass Covertype UCI dataset ($m=581012,\, n=54$) and the full Reuters RCV1 dataset ($m=804414,\, n=47236$) obtainable
from \url{http://www.jmlr.org/papers/volume5/lewis04a/lyrl2004_rcv1v2_README.htm}.
For the Covertype dataset we study the binary classification problem of the first class versus rest while for the RCV1 we consider both the binary text classification tasks of the C11 and CCAT classes versus rest. The Physics and Covertype datasets were rescaled by a multiplicative factor 0.001. The experiments were conducted on a 2.5 GHz Intel Core 2 Duo processor with 3 GB RAM running Windows Vista. Our codes written in C++ were compiled using the g++ compiler under Cygwin.

The parameter $\Delta$ of the extended space is chosen from the set $\{3,1,0.3,0.1\}$ in such a way that it corresponds approximately to $R/10$ or $R/3$ depending on the size of the dataset such that the ratio $\gamma_{\rm d} /R$ does not become too small (given that the extension generates a margin of at least $\Delta/\sqrt{m}$). More specifically, we have chosen $\Delta=3$ for Covertype, $\Delta=1$ for Adult, Web and Physics, $\Delta=0.3$ for Webspam, C11 and CCAT and $\Delta=0.1$ for Real-sim and News20. We also verified that smaller values of $\Delta$ do not lead to a significant decrease of the training error. For all datasets and for algorithms that introduce bias through augmentation the associated parameter $\rho$ was set to the value $\rho=1$.

We begin our experimental evaluation by comparing PDM with PFM. We run PDM with accuracy parameter $\epsilon=0.01$ and subsequently PFM with the fixed margin $\beta=(1-\epsilon)\gamma_{\rm d}$ set to the value $\gamma^{\prime}_{\rm d}$ of the directional margin achieved by PDM. This procedure is repeated using PDM-succ with step $\eta=8$ (i.e., $\epsilon_0=0.5, \epsilon_1=0.0625, \epsilon_2=\epsilon=0.01$). Our results (the value of the directional margin $\gamma^{\prime}_{\rm d}$ achieved, the number of required updates (upd) for convergence and the CPU time for training in seconds (s)) are presented in Table \ref{Table1}. We see that PDM is considerably faster than PFM as far as training time is concerned in spite of the fact that PFM needs much less updates for convergence. The successive run scenario succeeds, in accordance with our expectations, in reducing the number of updates to the level of the updates needed by PFM in order to achieve the same value of $\gamma^{\prime}_{\rm d}$ at the expense of an increased runtime. We believe that it is fair to say that PDM-succ with $\eta=8$ has the overall performance of PFM without the defect of the need for a priori knowledge of the value of $\gamma_{\rm d}$. We also notice that although the accuracy $\epsilon$ is set to the same value for both scenarios the margin achieved with successive runs is lower. This is an indication that PDM-succ obtains a better estimate of the maximum directional margin $\gamma_{\rm d}$. 
\begin{table}[t]
\centering
\caption{Results of an experimental evaluation comparing the algorithms PDM and PDM-succ with PFM.}
\label{Table1}
\begin{tabular}{|c|c|c|c|c|c||c|c|c|c|c|}
\hline\hline
\multicolumn{1}{|c|}{\ \makebox(3,3)[t]{data}  \ }&\multicolumn{3}{|c|}{\
PDM $\; \epsilon=0.01$ \ }& \multicolumn{2}{c||}{ \  PFM \ }  &  \multicolumn{3}{c|}{ \ PDM-succ $\; \epsilon=0.01$ \ } 
& \multicolumn{2}{c|}{ \ PFM \ } \\[0.5pt]
\cline{2-11}
set & $\sh{10^4}\gamma^{\prime}_{\rm d}$ & $\sh{10^{-6}}$upd & s & $\sh{10^{-6}}$upd  & s & $\sh{10^4}\gamma^{\prime}_{\rm d}$ & $\sh{10^{-6}}$upd & s & $\sh{10^{-6}}$upd & s \\
\hline
\shr{$Adult$} & 84.57 & 27.43 & 3.7 & 10.70 & 7.3      & 84.46 & 9.312 & 5.3 & 9.367 & 6.6 \\ 
\hline
\shr{$Web$} & 209.6 & 739.4 & 0.8 & 1.089 & 0.9        & 209.1 & 0.838 & 0.9 & 0.871 & 0.8  \\
\hline
\shr{$Physics$} & 44.54 & 9.449 & 10.4 & 6.021& 13.8    & 44.53 & 5.984 & 15.3 & 6.006 & 13.8 \\
\hline
\shr{$Real-sim$}  & 39.93 & 15.42 & 13.6 & 12.69 & 35.7 & 39.74 & 5.314 & 13.8 & 5.306 & 14.3  \\
\hline
\shr{$News20$}  & 91.90 & 2.403  & 27.4 & 1.060 & 55.6  & 91.68 & 0.814 & 47.7 & 0.813 &  43.7 \\
\hline
\shr{$Webspam$} & 10.05 & 331.0 & 197.5 & 108.4 & 348.0 & 10.03 & 89.72 & 247.0 & 89.60 & 264.5 \\
\hline
\shr{$Covertype$} & 47.51 & 189.7 & 86.6 & 68.86 & 156.0 & 47.48 & 66.03 & 146.1 & 64.41 & 142.5  \\
\hline
\shr{$C11$}  & 13.81 & 148.6 & 156.3& 75.26 & 895.1     & 13.77 & 49.02 & 612.4 & 49.22 & 557.5  \\
\hline
\shr{$CCAT$} & 9.279 & 307.7& 310.6 & 151.2 & 1923.5   & 9.253 & 107.8 & 1389.8 & 107.8 & 1601.0 \\
\hline
\end{tabular}
\end{table}

We also considered other large margin classifiers representing classes of algorithms such as perceptron-like algorithms, decomposition SVMs and linear SVMs with the additional requirement that the chosen algorithms need only specification of an accuracy parameter. From the class of perceptron-like algorithms we have chosen (aggressive) ROMMA
which is much faster than ALMA in the light of the results presented in \cite{IHT,PT1}. Decomposition SVMs are represented by ${\rm SVM}^{\rm light}$ \cite{Joa06} which, apart from being one of the fastest algorithms of this class, has the additional advantage of making very efficient use of memory, thereby making possible the training on very large datasets. Finally, from the more recent class of linear SVMs we have included in our study the dual coordinate descent (DCD) algorithm \cite{HCL} and the margin perceptron with unlearning (MPU)\footnote{MPU uses dual variables but is not formulated as an optimization. It is a perceptron incorporating a mechanism of reduction of possible contributions from ``very-well classified" patterns to the weight vector which is an essential ingredient of SVMs.} \cite{PT2}. We considered the DCD versions with 1-norm (DCD-L1) and 2-norm (DCD-L2) soft margin which for the same value of the accuracy parameter produce identical solutions if the penalty parameter is $C=\infty$ for DCD-L1 and $C=1/(2\Delta^2)$ for DCD-L2. The source for ${\rm SVM}^{\rm light}$ (version 6.02) is available at \url{http://smvlight.joachims.org} and for DCD at \url{http://www.csie.ntu.edu.tw/~cjlin/liblinear}. The absence of publicly available implementations for ROMMA necessitated the writing of our own code in C++ employing the mechanism of active sets proposed in \cite{IHT} and incorporating a mechanism of permutations performed at the beginning of a full epoch. For MPU the implementation followed closely \cite{PT2} with active set parameters $\bar c=1.01$, $N_{\rm ep_1}=N_{\rm ep_2}=5$, gap parameter $\delta b=3R^2$ and early stopping.  

\begin{table}[t]
\caption{Results of experiments with ROMMA, ${\rm SVM}^{\rm light}$, DCD-L1, DCD-L2 and MPU algorithms. The accuracy parameter for all algorithms is set to 0.01.}
\label{Table2}
\centering
\begin{tabular}{|c|c|c|c|c|c|c|c|c|c|} 
\hline\hline
\multicolumn{1}{|c|}{\ \hspace{4pt} \mbox{\makebox(3,3)[t]{data}}  \hspace{4pt} \ }  & \multicolumn{2}{c|}{\  \hspace{8pt} ROMMA \hspace{8pt} \  } &  \multicolumn{2}{c|}{\  \hspace{8pt} ${\rm SVM}^{\rm light}$ \hspace{8pt} \  } &  
\multicolumn{2}{c|}{\  \hspace{3pt} DCD-L1  \hspace{8pt} \  } & 
\multicolumn{1}{c|}{DCD-L2} & \multicolumn{2}{c|}{\  \hspace{8pt}${\rm MPU}$  \hspace{8pt} \  }\\[0.5pt]
\cline{2-10}

 \mbox{\makebox(10,8)[t]{set}} & \mbox{\makebox(10,12)[b]{$\sh{10^4} \gamma^{\prime}_{\rm d}$}} &  \mbox{\makebox(10,8)[c]{s}} & \mbox{\makebox(10,12)[b]{$\sh{10^4} \gamma^{\prime}$}} & \mbox{\makebox(10,8)[c]{s}} &   \mbox{\makebox(10,12)[b]{$\sh{10^4} \gamma^{\prime}_{\rm d}$}}  & \mbox{\makebox(10,8)[c]{s}} & \mbox{\makebox(10,8)[c]{s}} & \mbox{\makebox(10,12)[b]{$\sh{10^4} \gamma^{\prime}_{\rm d}$}} &  \mbox{\makebox(10,8)[c]{s}}  \\
\hline
\shr{$Adult$} & 84.66 & 275.8   & 84.90  & 414.2  & 84.95   &  0.6 & 0.5 & 84.61 & 0.8 \\ 
\hline
\shr{$Web$}  & 209.6  & 52.6    & 209.4  & 40.3   & 209.5   &  0.7 & 0.6 & 209.5 & 0.3 \\
\hline
\shr{$Physics$}&44.57 & 117.7   &  44.60 & 2341.8 & 44.57   &  22.5 & 20.0 & 44.62 & 4.9 \\
\hline
\shr{$Real-sim$}& 39.89&1318.8  & 39.80  & 146.5  & 39.81   &  6.4  & 5.6  & 39.78 & 3.3 \\
\hline
\shr{$News20$} & 92.01 & 4754.0 & 91.95  & 113.8  & 92.17   &  48.1 & 47.1 & 91.62 & 15.8 \\
\hline
\shr{$Webspam$}& 10.06 &39760.6 & 10.07  & 29219.4& 10.08   &  37.5 & 33.0 & 10.06 & 28.2 \\
\hline
\shr{$Covertype$}&47.54& 43282.0& 47.73  & 48460.3& 47.71   &  18.1 & 15.0 & 47.67 & 18.7 \\
\hline
\shr{$C11$}    & 13.82 &146529.2& 13.82  & 20127.8& 13.83   &  30.7 & 27.2 & 13.79 & 20.2 \\
\hline
\shr{$CCAT$}  & 9.290 & 298159.4& 9.291  & 83302.4& 9.303   &  51.9 & 46.2 & 9.264 & 36.1 \\
\hline
\end{tabular}
\end{table} 
The experimental results (margin values achieved and training runtimes) involving the above algorithms with the accuracy parameter set to 0.01 for all of them are summarized in Table \ref{Table2}. Notice that for ${\rm SVM}^{\rm light}$ we give the geometric margin $\gamma^{\prime}$ instead of the directional one $\gamma^{\prime}_{\rm d}$ because ${\rm SVM}^{\rm light}$ does not introduce bias through augmentation. For the rest of the algorithms considered, including PDM and PFM, the geometric margin $\gamma^{\prime}$ achieved is not listed in the tables since it is very close to the directional margin $\gamma^{\prime}_{\rm d}$ if the augmentation parameter $\rho$ is set to the value $\rho=1$. Moreover, for DCD-L1 and DCD-L2 the margin values coincide as we pointed out earlier. From Table \ref{Table2} it is apparent that ROMMA and ${\rm SVM}^{\rm light}$ are orders of magnitude slower than DCD and MPU. Comparing the results of Table \ref{Table1} with those of Table \ref{Table2} we see that PDM is orders of magnitude faster than ROMMA which is its natural competitor since they both belong to the class of perceptron-like algorithms. PDM is also much faster than ${\rm SVM}^{\rm light}$ but statistically a few times slower than DCD, especially for the larger datasets. Moreover, PDM is a few times slower than MPU for all datasets. Finally, we observe that the accuracy achieved by PDM is, in general, closer to the before-run accuracy 0.01 since in most cases PDM obtains lower margin values. This indicates that PDM succeeds in obtaining a better estimate of the maximum margin than the remaining algorithms with the possible exception of MPU. 

Before we conclude our comparative study it is fair to point out that PDM is not the fastest perceptron-like large margin classifier. From the results of \cite{PT1} the fastest algorithm of this class is the margitron which has strong before-run guarantees and a very good after-run estimate of the achieved accuracy through (\ref{est}). However, its drawback is that an approximate knowledge of the value of $\gamma_{\rm d}$ (preferably an upper bound) is required in order to fix the parameter controlling the margin threshold. Although there is a procedure to obtain this information, taking all the facts into account the employment of PDM seems preferable.    

\section{Conclusions}
We introduced the perceptron with dynamic margin (PDM), a new approximate maximum margin classifier employing the classical perceptron update, demonstrated its convergence in a finite number of steps and derived an upper bound on them. PDM uses the required accuracy as the only input parameter. Moreover, it is a strictly online algorithm in the sense that it decides whether to perform an update taking into account only its current state and irrespective of whether the pattern presented to it has been encountered before in the process of cycling repeatedly through the dataset. This certainly does not hold for linear SVMs. Our experimental results indicate that PDM is the fastest large margin classifier enjoying the above two very desirable properties.

\end{document}